%% file: nic-buw-ki-paper.tex
\documentclass{nic-series}

\usepackage{xcolor}
\usepackage{subfig}

\usepackage{wrapfig}
\usepackage{cleveref}

\usepackage{tikz}
\usetikzlibrary{shapes}\usepackage{pgfplots}
\pgfplotsset{compat=newest}
\usepackage{graphicx}
\usepackage{marvosym}

\begin{document}

\title{Uncertainty Quantification and Resource-Demanding Computer Vision Applications of Deep Learning}

\author{Julian~Burghoff \inst{1} \and
Robin~Chan \inst{1} \and
Hanno~Gottschalk \inst{1} \and 
Annika~Mütze \inst{1} \and
Tobias~Riedlinger \inst{1} \and
Matthias~Rottmann \inst{1} \and
Marius~Schubert \inst{1}}

\authortoc{J.~Burghoff\inst{1}, R.~Chan\inst{1}, H.~Gottschalk\inst{1}, A.~Mütze\inst{1},  T.~Riedlinger\inst{1},  M.~Rottmann\inst{1}, M.~Schubert\inst{1}}

\institute{School of Mathematics \& Natural Sciences, IMACM, IZMD\\
         University of Wuppertal, D-42119 Wuppertal, Germany\\
         \email{\{burghoff, rchan, hanno.gottschalk, muetze, riedlinger, rottmann, mschubert\}@uni-wuppertal.de}
          }

\maketitle

\begin{abstracts}
Bringing deep neural networks (DNNs) into safety critical applications such as automated driving, medical imaging and finance, requires a thorough treatment of the model's uncertainties. Training deep neural networks is already resource demanding and so is also their uncertainty quantification. In this overview article, we survey methods that we developed to teach DNNs to be uncertain when they encounter new object classes. Additionally, we present training methods to learn from only a few labels with help of uncertainty quantification. Note that this is typically paid with a massive overhead in computation overhead of an order of magnitude and more compared to ordinary network training. Finally, we survey our work on neural architecture search which is also an order of magnitude more resource demanding then ordinary network training.
\end{abstracts}

\section{Introduction}

Deep learning (DL), convolutional neural networks in particular, in the last 20 years has revolutionised computer vision \cite{lecun1990handwritten,goodfellow2016deep}. This has brought new applications into reach that previously have been considered impossible. The application of deep learning technology in safety critical fields -- with autonomous driving or assistance of medical diagnosis being prominent examples -- however poses new challenges, as deep learning is a statistical machine which errors with a certain probability. Additional information on the reliability of the prediction of AI algorithms should therefore be considered as an integral part of the model's inference. This will either be a piece of valuable information that is provided by the AI-driven assistance to a human decision maker, like in medicine, or to downstream algorithms that, in the case of autonomous driving, choose a more defensive driving policy when facing high uncertainty. 

Uncertainty quantification (UQ) for deep neural networks has been under intense study in the past years. The classification of probabilistic uncertainty -- epistemic, aleatoric -- has been analysed, e.g., using Bayesian neural networks\cite{doya2007bayesian} or Monte Carlo dropout\cite{gal2016theoretically} as a proxy for Bayesian epistemic uncertainty. For an introduction to UQ, we refer to a survey\cite{hullermeier2019aleatoric}. UQ is also needed in open worlds problems: As the semantics of AI based classifiers only contains from few to several dozen categories, unknown objects outside this closed semantic world are forcibly classified as an instance of one of the known categories. Control and traceability in such scenarios therefore requires an adequate expression of uncertainty by the network \cite{meinke2019towards,papadopoulos2019outlier,hendrycks2016baseline} or even abstention \cite{nguyen2019reliable,cortes2016boosting}.

Advanced deep learning models enable the task of object detection and localisation: on a high resolution image, multiple objects are detected \cite{redmon2016you,redmon2018yolov3} or the entire image is segmented into semantic categories providing a pixel-wise classification, see e.g.\ \cite{he2016deep,chollet2017xception,sandler2018mobilenetv2}. The training of such models requires heavy computer resources, in particular general purpose graphic processing units (GPGPU) \cite{chollet2017xception}. Combining such advanced prediction models with an adequate UQ therefore is an important task.  

In the past years, UQ for image segmentation and object detection has been initiated in a series of works \cite{rottmann2018,Rottmann2019,Schubert2019,Maag2019,schubert2020metadetect,riedlinger2021gradient} focusing on false positive instances and false negative instances \cite{chan2019application,chan2019dilemma}, respectively, we also refer to a survey \cite{Rottmann2019}. While many works focus on street scene recognition for autonomous driving, \cite{rottmann2018,mehrtash2019confidence} cover medical image segmentation applications as well.

Further application of UQ is the usage in active \cite{mackowiak2018cereals,colling2020metabox} and semi-supervised training strategies \cite{doya2007bayesian}, which is eligible to reduce the required ground truth by almost an order of magnitude. 
Labelling real-world data is time consuming, costly and error-prone \cite{coco}. 
To mitigate these problems several simulation tools and synthetic datasets \cite{Dosovitskiy17,richter_playing_2016,synthia} were published in the recent years. 
Using this data mitigates the label effort but leads to uncertainty of the network as it predicts on a different domain than it was trained on. 
Reducing via so called domain adaptation methods and analysing the uncertainty is an active research field.
It has also been demonstrated that network architectures that were pruned can be more robust \cite{Guo18}.
The UQ and the training of deep networks for uncertainty awareness however require additional hardware resources compared with the already resource hungry standard deep learning technology, which requires plenty of GPU and CPU compute resources. 

In the remainder of this work, we give an overview of current projects of the authors that use GPU compute resources on the JUWELS\cite{JUWELS} supercomputer at the Jülich Super Computing Centre (JSC). We present insights into results and discuss open questions. More precisely, the remainder of this work is structured as follows:
In \cref{sec:robin}, we present a method to identify objects from semantically unknown categories. Intuitively, these objects should come with high model prediction uncertainty, which we exploit to detect and localise unknown objects in semantic segmentation.
In \cref{sec:annika} we analyse the domain gap as a source of uncertainty of a neural network. In the context of urban street scenes and simulations we present a semi-supervised domain adaptation approach for the semantic segmentation task.
In \cref{sec:marius und tobi}, we describe our active learning endeavours in the context of deep object detection.
We compare a method based on previously developed, highly informative uncertainty quantification methods against common baseline approaches.
\Cref{sec:julian} is about the automated optimisation of neural networks with respect to their architecture. We show how we have extended Google's MorphNet approach to build networks from scratch. This is demonstrated in numerical experiments on the CIFAR10 dataset.







\section{Detecting Unknown Objects in Semantic Segmentation} \label{sec:robin}

\begin{figure}[t]
    \centering
    \subfloat[Semantic segmentation mask]{\includegraphics[width=0.45\linewidth]{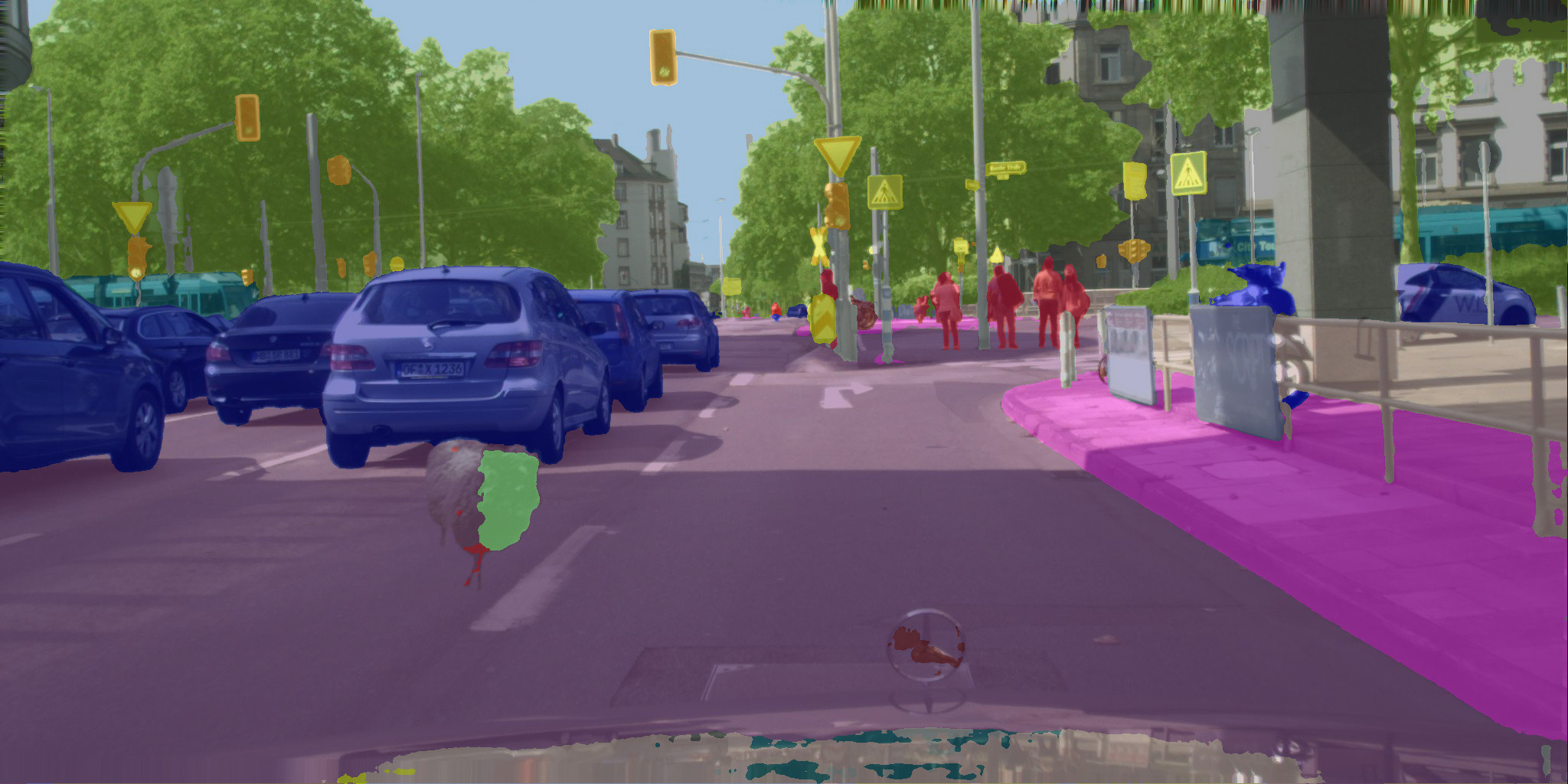}} ~
    \subfloat[Pixel-wise softmax entropy scores]{\includegraphics[width=0.45\linewidth]{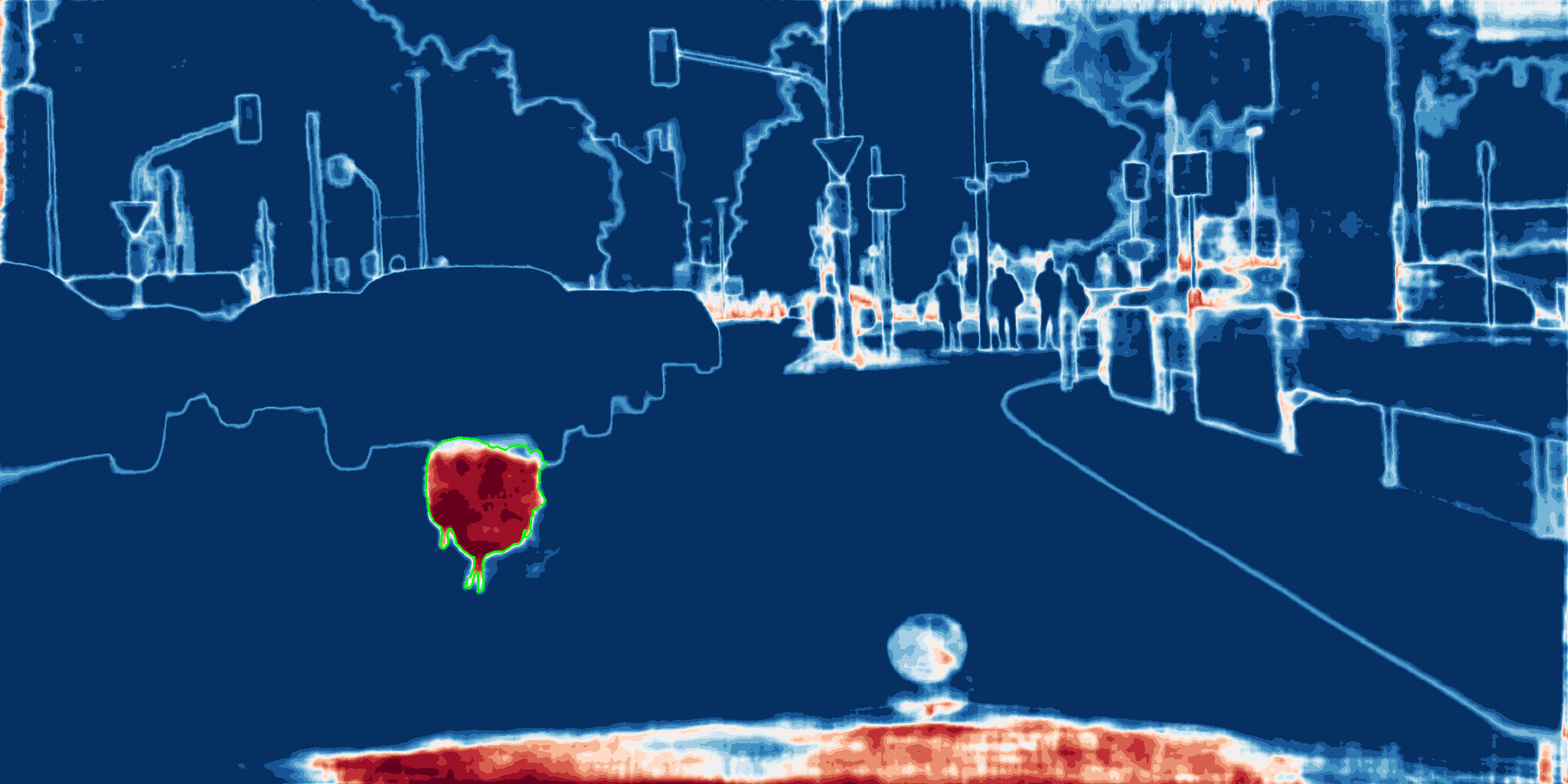}}
    \caption{Illustration of the successful detection of an unknown object by means of the softmax entropy in semantic segmentation.
    }
    \label{fig:ood-segmentation}
\end{figure}

Semantic segmentation is the computer vision task of assigning each pixel in a given image to an object category. For this complex task, one typically employs very deep convolutional neural networks (DNNs). Their training typically requires plenty of GPU resources. Depending on the architecture of the DNN, such a training can take up to one week on a single GPU. DNNs are trained to operate on a closed and pre-defined set of object classes. Therefore, those models are ill-equipped to handle objects from classes that are semantically unknown. These outlined objects are also called \emph{out-of-distribution} (OOD), as they are associated with an extremely low probability of occurrence. We introduce a method to detect such unknown objects in semantic segmentation \cite{Chan2020entropy}, which is commonly known as the task of \emph{anomaly segmentation}. Our proposed training approach enforces high prediction uncertainty on objects from unknown categories, while at the same time retaining segmentation performance on original categories. Our approach outperforms several anomaly segmentation baselines, such as Mahalanobis distance \cite{Lee2018mahala} or Monte Carlo dropout \cite{gal2016}. We achieve significant anomaly segmentation performance gains over those established baselines of up to $26$ percent points in the relevant metric area under precision recall curve, yielding one of the best performing methods on the public benchmarks Fishyscapes \cite{Blum19fishyscapes} and SegmentMeIfYouCan \cite{chan2021segmentmeifyoucan} at the time of writing. For an illustration of our method we refer to \Cref{fig:ood-segmentation}.
For finding the appropriate training hyperparameters of our anomaly segmentation model but also other parameters such as the size of the OoD dataset and OoD weighting in the multi-criteria loss function, extensive computational experiments were necessary. To this end, we used single-GPU training on single (accelerated) nodes performing four runs simultaneously. In each run a semantic segmentation DNN with more than 60 million weight parameters was trained for multiple epochs.

\begin{figure}[t]
    \centering
    \includegraphics[width=.75\linewidth]{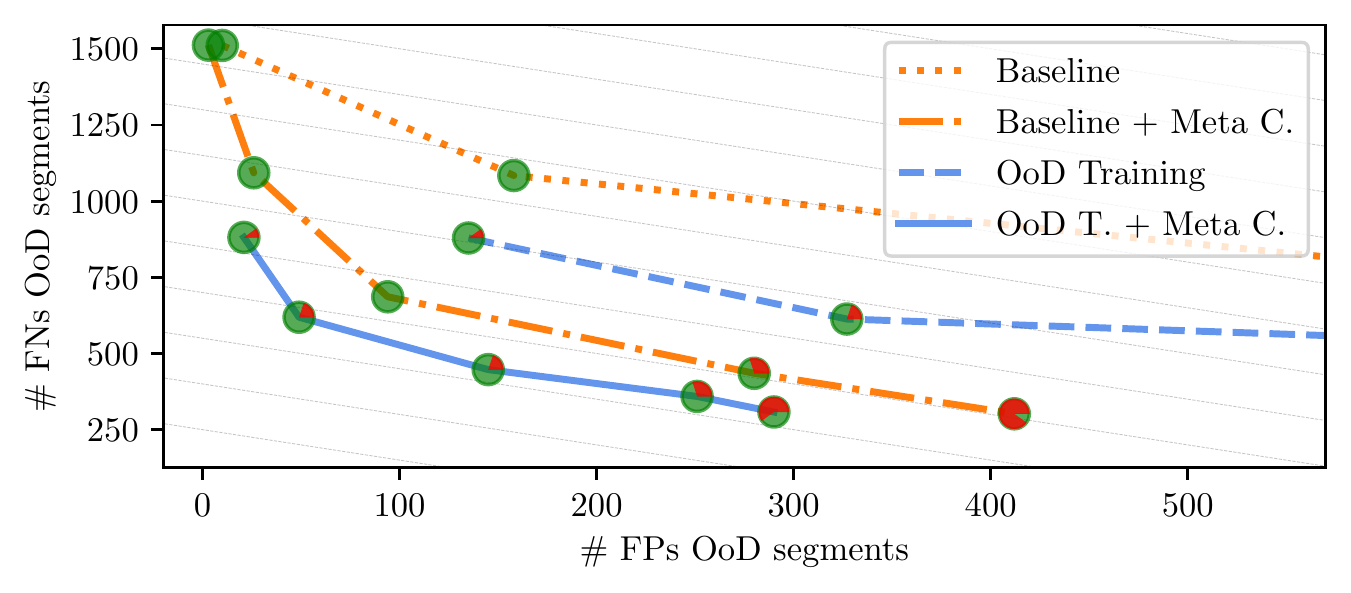}
    \caption{Detection errors of OoD objects. In this plot, the number of false positive and false negative errors are plotted against each other for different combinations of entropy maximisation (denoted as OOD Training) and meta classification. For each combination the detection threshold is additionally varied, i.e., the entropy value at which an object prediction (or segment in the case of semantic segmentation) is identified as OOD. The corresponding pie-chart markers indicate the loss in original semantic segmentation performance, being entirely red if this loss is greater than 1\%.}
    \label{fig:error_rates}
\end{figure}

Furthermore, we introduce meta classification for anomaly segmentation. A meta classifier is a model that flags incorrect predictions. In this work, we employ a logistic regression as meta classifier to detect false indications of unknown objects based on hand-crafted metrics derived from the probabilistic output of the semantic segmentation DNN.
In our experiments, the additional usage of meta classification consistently reduces the number of false OoD indications, i.e.\,false positives \cite{chan2019metafusion}. Compared to the baseline, which is no entropy maximisation and no meta classification, we reduce the number of false positives by 66\% while at the same time reducing the number false negatives (overlooked OoD objects) by 50\%. Importantly, this significant improvement in the error rates is associated with only a marginal loss in original semantic segmentation performance of at most 1\%, see also \Cref{fig:error_rates}.
Moreover, as logistic regression models are simple linear models, it allows us to track features contributing to the detection of unknown objects. Therefore, besides observing an additional gain in anomaly segmentation performance the combination of entropy maximisation and meta classification particularly contributes to safer DNNs for real-world applications.

\section{Downstream Task informed Domain Adaptation GAN for Semantic Segmentation} \label{sec:annika}
As introduced in the previous section, deep neural networks (DNNs) are used for semantic segmentation tasks. For generalising well on unseen scenes, these models need plenty of labelled images but a manual label process is time and cost consuming and usually error-prone.
With the help of simulations we have the possibility of generating arbitrarily many labelled data samples to train an expert network on the simulated domain via supervised learning. However, DNNs trained on one domain can perform arbitrarily poor when switching to another domain, i.e., when changing the data generating distribution \cite{csurka_domain_2017}.
With our approach we try to mitigate the so-called domain gap via style transfer and guidance towards the down stream task on the domain where labelled data is rare. Our work is based on the publication of \cite{CycleGAN2017} and is extended by the semi-supervised training routine of the GAN and an additional loss term to achieve a task aware generator. 
Our method can be split into three steps:
\begin{itemize}
    \item \textit{Semantic Segmentation:} Training a semantic segmentation network on the synthetic domain (expert network);
    \item \textit{Pre-training:} Transfer the real images into the synthetic domain via Image-to-Image translation;
    \item \textit{Fine-tuning:} Guide the generator to the downstream task with the help of a couple of ground truth masks and the semantic segmentation loss.
\end{itemize}
Training neural networks and the style transfer in particular on images with a high resolution is computationally expensive and takes about 45 minutes per epoch on a single NVIDIA V100 GPU each. As the network training and the GAN pre-training are independent, we can train both in parallel on different GPUs.
For our experiments, we focus on the task of semantic segmentation for real world street scenes where we have few labels and use a simulation to train our expert segmentation network. Therefore we focus on ``real to sim(ulation)'' as domain shift. For now we consider random sampling as query strategy for the fine tune process.
We train the networks completely from scratch to prevent a bias towards the real world, to evaluate the domain gap accurately. As a consequence we accept a reduction of the total accuracy.
The Cityscapes dataset \cite{cityscapes} serves as real domain in our experiments and we either use Synthia\_Rand\_Cityscapes \cite{synthia} or a self generated CARLA dataset \cite{Dosovitskiy17} as synthetic domain. Our experiments show that even though our overall performance lies between $28$--$39\%$ mean Intersection over Union (mIoU) we can mitigate the domain gap. We vary the amount of labelled data in our experiments between $0.5\%$ ($5$ images) to $10\%$ ($297$ images) of the Cityscapes trainingset. We achieve a gap reduction by $18$--$30$ percent points mIoU depending on the amount of labels used during fine-tuning. 
\begin{figure*}[t!]
    \centering
    \includegraphics{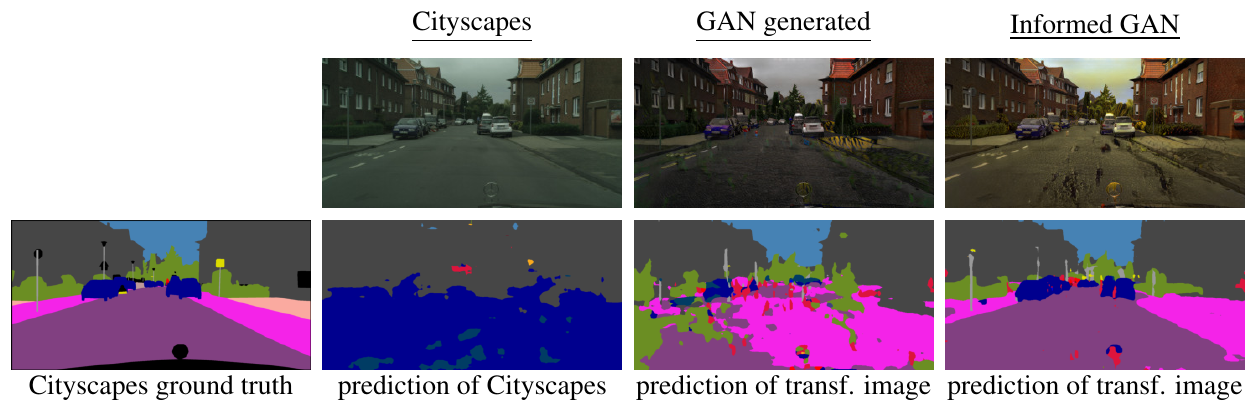}
    \caption{Comparison of prediction results of original Cityscapes image (left), simple style transfer (mid) and our approach (right).
    \vspace{-4ex}}
    \label{fig:pred_comparison} 
\end{figure*}
\begin{figure}[t]
    \centering
        \subfloat[Network trained on Synthia]{\includegraphics[width=0.45\textwidth]{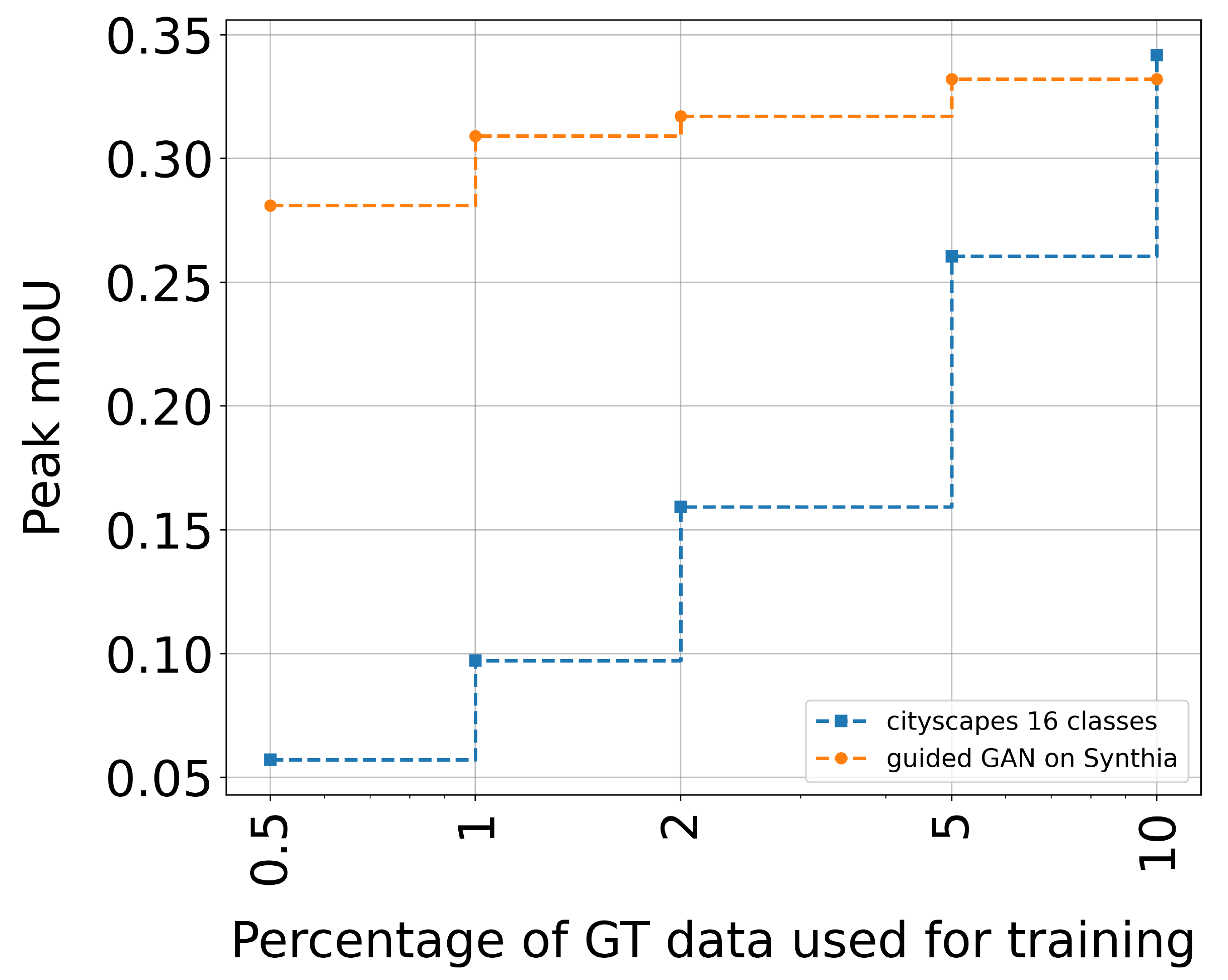}}
    \hfill
    \subfloat[Network trained on CARLA]{
        \includegraphics[width=0.45\textwidth]{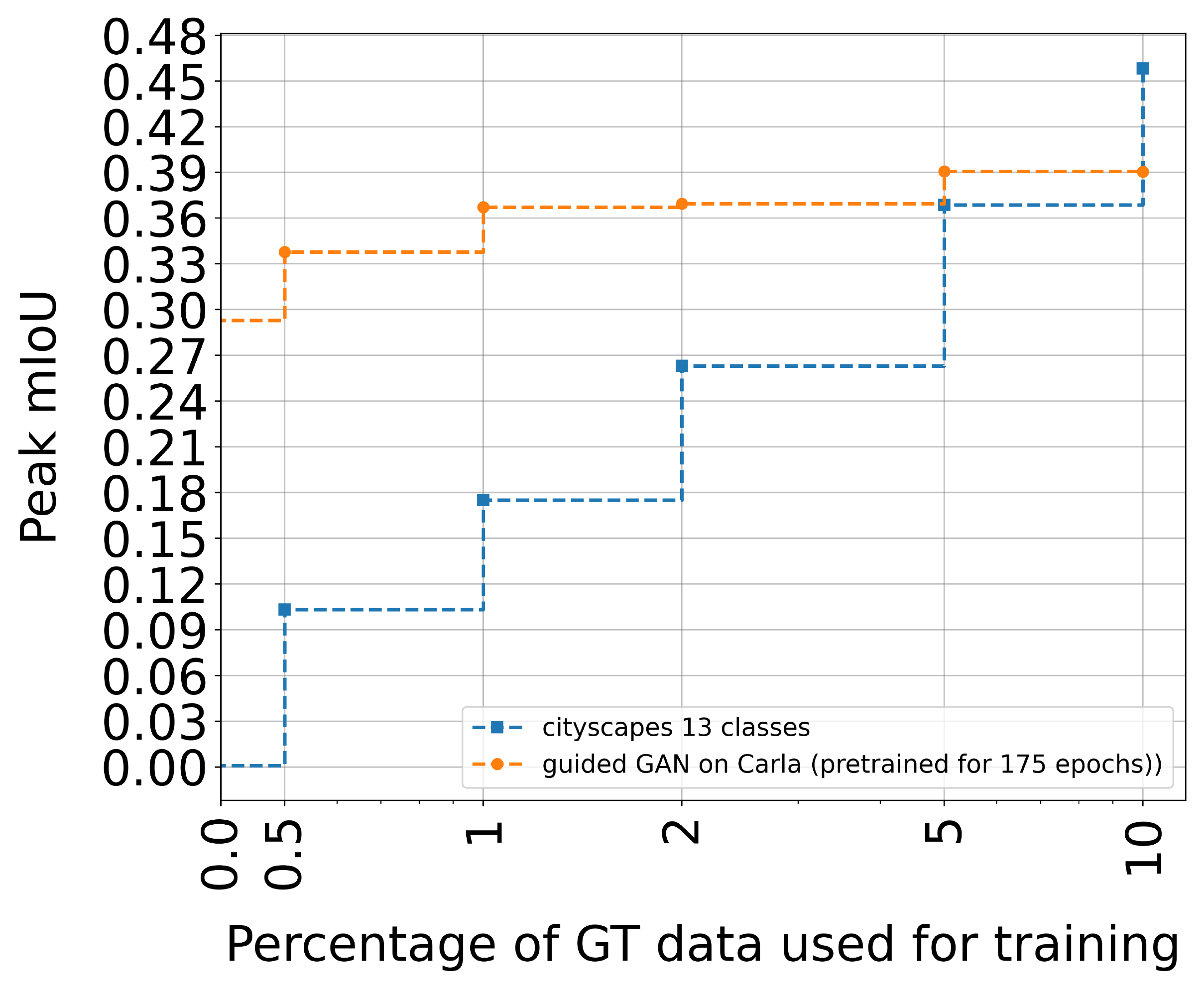}}
    \caption{Comparison of the performance given different amount of ground truth data. The blue graph represents the supervised Cityscapes training and the orange graph our approach. Please note the different scaling of the y-axis in the two plots.
    \vspace{-3.5ex}}
    \label{fig:data_vs_miou}
\end{figure}

An example image from the validation set of Cityscapes is shown in \cref{fig:pred_comparison}. For a qualitative comparison we show its transformation with plain CycleGAN \cite{CycleGAN2017} and with an informed GAN which was trained with our approach and $148$ randomly sampled images with labels during the fine tuning step. Furthemore the corresponding prediction of the network trained on Synthia is illustrated in the second row. One can see, that even though we still find some artifacts in the transformed images, an improvement in the overall scene understanding.
In addition we compare our approach to a plain supervised Cityscapes training with the same amount of labels as used for the fine tuned. In \cref{fig:data_vs_miou} the comparison of the strongest performance achieved for a given amount of ground truth (GT) is shown. On the x-axis the relative amount of the Cityscapes training data set (2975 images) is plotted. The plot shows that with our approach we can improve segmentation results when only a small fraction of labelled data is available. But it is worth noting, that if the amount of labelled data increases, a supervised training approach should be preferred as the information of the labels can be learned directly.

\section{Active Learning Strategies for Deep Object Detection} \label{sec:marius und tobi}
It is widely known that, as a rule of thumb, for the training of any kind of artificial neural network, an increase of training data leads to increased performance and better generalisation.
Obtaining labelled training data, however, is often very costly since in most cases, labels can only be given by a human ``oracle'' / expert.
In applications where labelled training data is scarce (such as in deep object detection), the application of active learning is promising for an effective cost reduction with respect to training labels. 
The generic active learning algorithm works as follows: after training a network on some (potentially rather small) initial set of labelled data $\mathcal{L}$ to convergence, the network's uncertainty with respect to samples from some unlabelled set of data $\mathcal{U}$ is inferred.
It has been shown in \cite{gal,settles.tr09} that in many cases, labels corresponding to uncertain samples can increase the network's performance significantly when adding those samples to the training data.
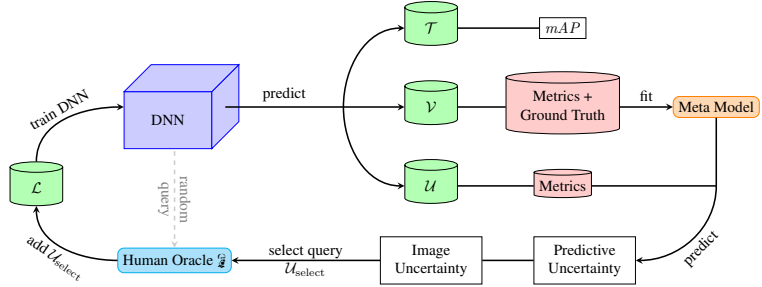
\begin{figure}
    \centering
    \resizebox{0.8\textwidth}{!}{
    \input{images/meta_detect_pipeline.tex}
    }
    \caption{Schematic setup of active leaning cycle for MetaDetect-based query strategies.}
    \label{fig:metadetect_al_cycle}
\end{figure}

\begin{figure}
    \centering
    \resizebox{0.8\textwidth}{!}{
    \input{images/al_nutshell.tex}
    }
    \caption{Semi-synthetic dataset generation for MNIST digit and EMNIST letter detection on a complex background.}
    \label{fig:nutshell_generation}
\end{figure}
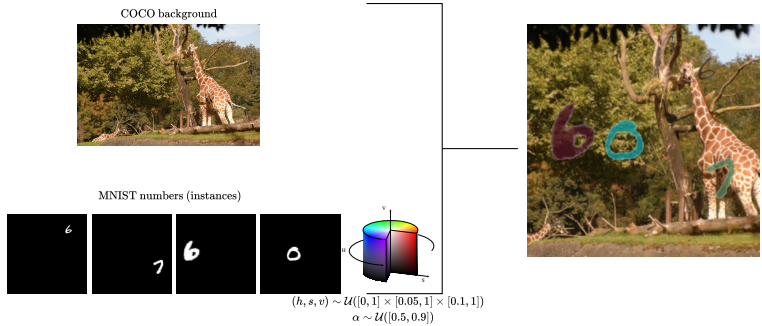

Our proposed active learning strategy builds upon previous work in uncertainty quantification for deep object detection\cite{schubert2020metadetect, riedlinger2021gradient}.
The MetaDetect \cite{schubert2020metadetect} and gradient uncertainty \cite{riedlinger2021gradient} approaches have shown state-of-the-art uncertainty quantification in terms of false positive detection of individual predicted instances as well, as localisation uncertainty estimation.
Moreover, the proposed methods do not affect the training process of the neural network, therefore, fitting as a post-processing module fitted on a validation dataset $\mathcal{V}$ on top of any object detector making the implementation into an active learning cycle flexible.
\Cref{fig:metadetect_al_cycle} shows the experimental setup of this strategy starting with the initially labelled dataset $\mathcal{L}$ on the very left of the schematics.
\begin{figure}
    \centering
    \includegraphics[width=0.45\textwidth]{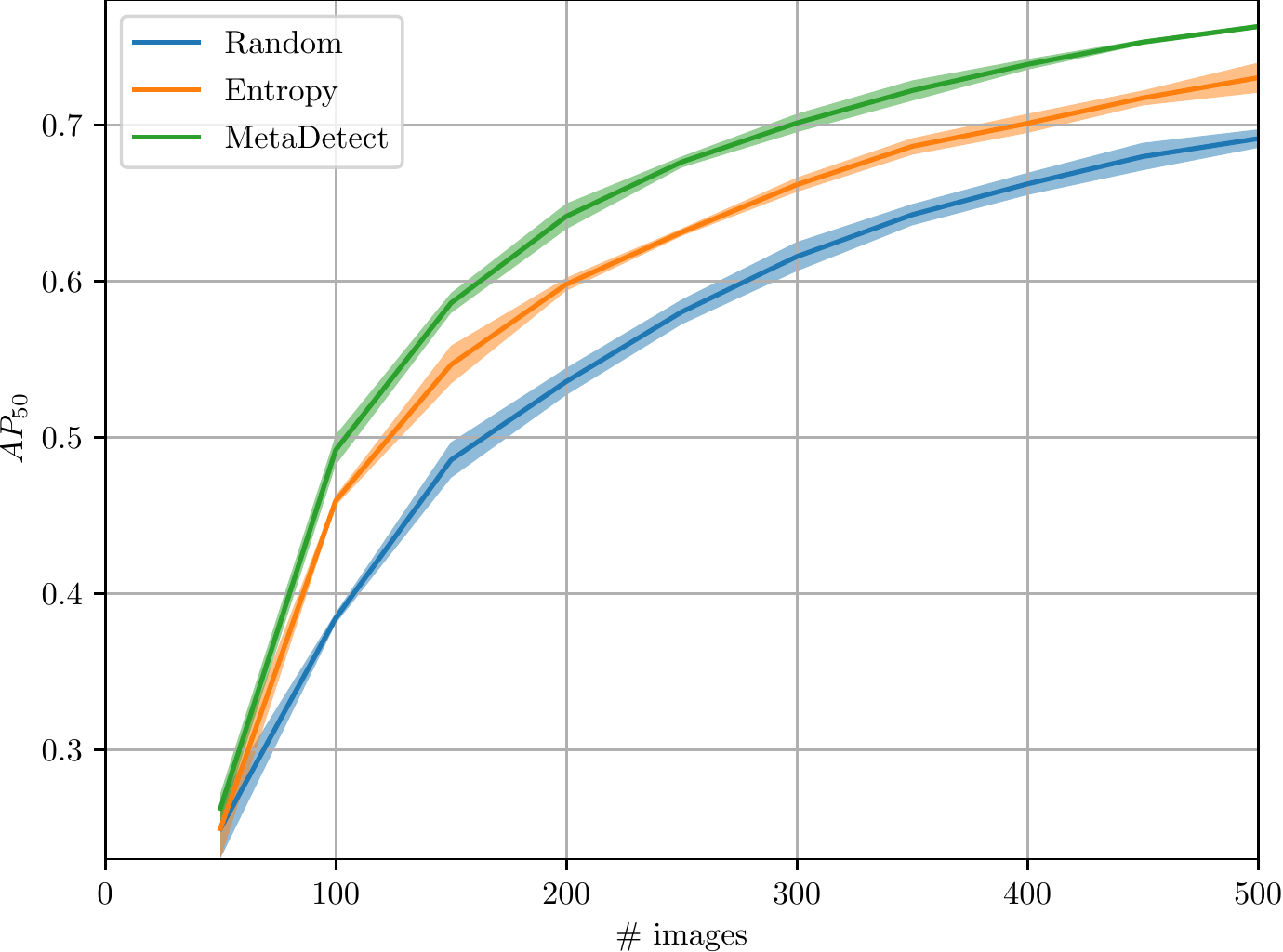}
    \includegraphics[width=0.5\textwidth]{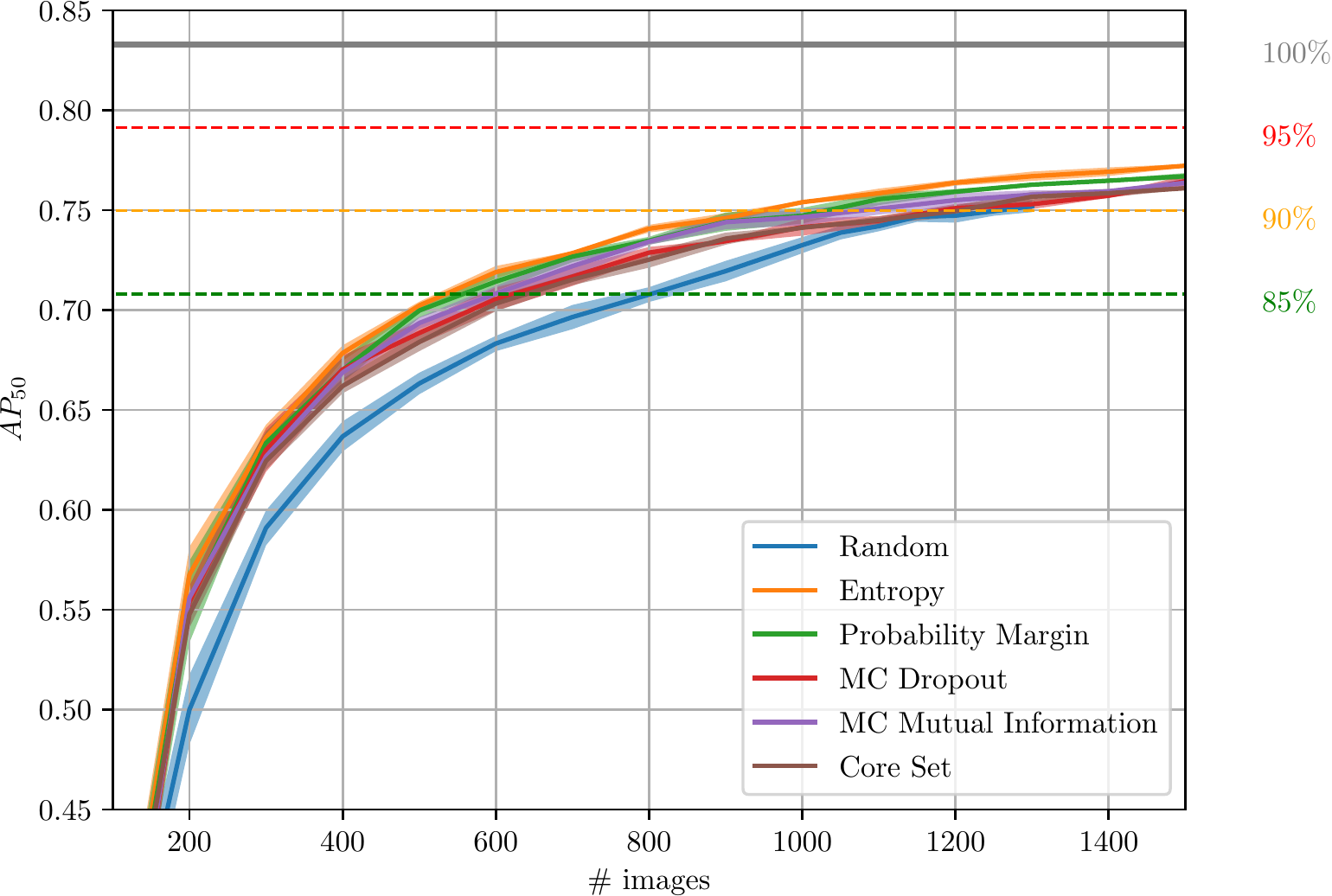}
    \caption{\emph{Left}: MetaDetect-based query strategy on our MNIST-based toy model in comparison with the random and the entropy baseline ($y$-axis shows detection performance in terms of $\mathit{AP}_{50}$).
    \emph{Right}: Comparison of common literature baselines on our EMNIST-based toy model.
    }
    \label{fig:metadetect_vs_baselines}
\end{figure}

Applying active learning to tasks that are close to real-world applications, such as object detection, requires large amounts of computational resources.
In each single experiment (e.g., 10-15 active learning cycles/query steps), a deep neural network for object detection needs to be trained until convergence several times which comes with a large computational overhead, even when utilising GPUs.
Apart from comparing to various baselines, in order to obtain reliable studies, each experiment needs to be repeated under different random seeds for upward of four times.
Moreover, contemporary active learning literature for object detection tends to use setups widely differing in the utilised hyperparameters.
When developing a query strategy and comparing to different baseline approaches, a rapid prototyping frequency is essential for the reduction of overall compute expense.
We, therefore,  base our investigation on an efficient, comparable and trackable toy model environment, in addition to standard setups using standard object detection benchmarks and networks.
Our toy model framework consists of two semi-synthetic data sets for supervised learning of detecting MNIST digits or, respectively, EMNIST letters on an MS COCO background image, see \cref{fig:nutshell_generation}.
Moreover, we use a slim network architecture in RetinaNet with a short ResNet18 backbone for fast training and inference, combined with the data sets cutting the time used for an active learning step (one cycle in \cref{fig:metadetect_al_cycle}) to up to 10\% of the time consumed in the standard setting.
This leads to an effective time consumption of up to 1 day per experiment as opposed to 5 to 10 days (when using a single GPU).
However, even this experimental setup still requires a considerable amount of GPU resources.

Central to our prototyping investigation is the comparison to a wide range of baselines beyond the random image selection and entropy-based query strategies which is computationally expensive despite the use of an efficient framework.
Our considerations lie on baselines which are not heavily architecture-dependent like the probability margin of the classifier, Monte-Carlo dropout-based selection strategies or the core set approach which was initially introduced for the classification setting.
See the right-hand side of \cref{fig:metadetect_vs_baselines} for a comparison of the implemented baselines.
We see that all baselines show significant improvement in comparison with the random strategy which represents uninformed image selection.


\begin{figure}
    \centering
    \includegraphics[width=0.99\linewidth]{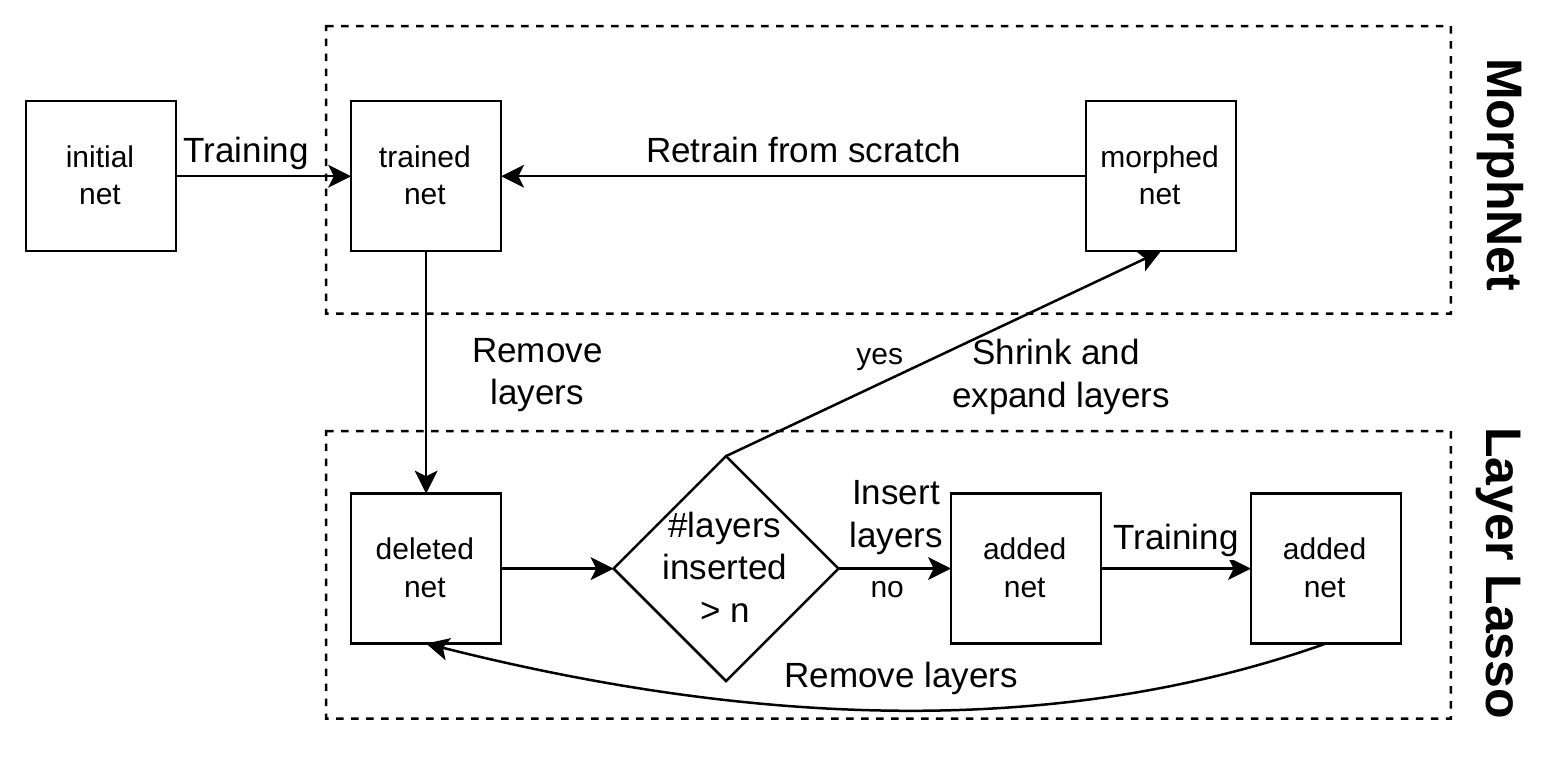}
    \captionsetup{width=\linewidth}
    \captionof{figure}{Training pipeline for generating a well-fitted architecture for a specific dataset. Starting with an initial architecture (which can consist of only one convolutional layer or can be a stat-of-the-art architecture) we train this net from scratch and then apply the Layer Lasso procedure. This adds and deletes residual blocks of convolutional layers according to the importance of the layer in terms of the sum of the absolute values of the weights. After a chosen number of Layer Lasso iterations $n$, the MorphNet procedure starts where the number of channels in each layer is shrunk (dependant on the sum of the weights of the filter) and expanded by a linear factor again to take use of the full computation capacity. At the end of the MorphNet routine the net is trained from scratch and the Layer Lasso method starts again with $0$ layers inserted.}
    \label{fig:trainingspipeline_autoML}
\end{figure}

\section{Automated Design of Neural Network Architectures} \label{sec:julian}
Since a neural network usually delivers high performance only in a small, predefined area of application (e.g.\ image recognition for a given set of classes), research is being conducted into how suitable architectures of neural networks can be found automatically.
To this end, we developed an architecture search method for convolutional neural networks, which is based on Google's MorphNet \cite{Gordon17}. Our extension for the first time allows to adaptively add and remove layers as well as whole residual blocks \cite{he2016deep} alongside with MorphNet, enabling users to automatically obtain neural networks architectures with close to state-of-the-art performance without any manual architecture development. This is particularly useful in exotic applications and/or new datasets where there are no readily trained networks available.

\begin{figure}
    \centering
    \includegraphics[width=.8\linewidth]{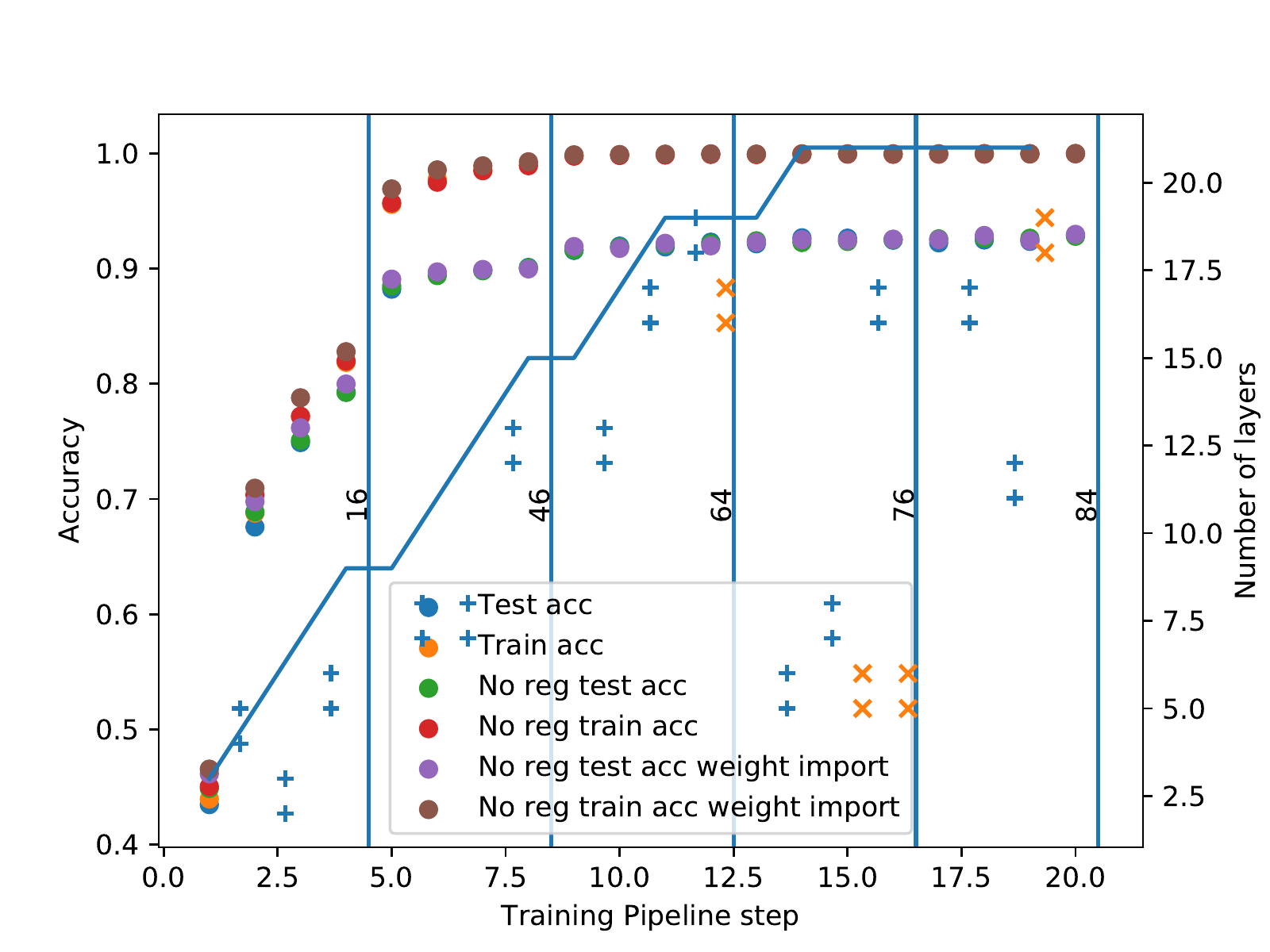}
    \captionof{figure}{Example run on the CIFAR10 dataset. Each pipeline step stands for a different tested architecture whereas points show the reached accuracy of this architecture in three different ways: Blue/orange points show the accuracy achieved by the network when all regularization terms (L2 + MorphNet + Layer Lasso) apply. Green and red points show the accuracy after training the same architecture from scratch but only with L2 regularization and violet/brown points are also trained only with L2 regularization, but we import the weights we achieved from the run with all regularization terms. The vertical blue lines indicate that a morphnet steps happens and the number next to the lines explain how many channels the initial convolutional layer, i.e.\ also all end layers of the different residual blocks, have. A blue plus/orange $x$ show at which position a layer was added/deleted, respectively.}
    \label{fig:CIFAR10_example}
\end{figure}

\Cref{fig:trainingspipeline_autoML} depicts the architecture search pipeline which is used to automatically search for new architectures of neural networks fitted to a given dataset. Besides the ability of adding and deleting blocks of layers in a similar way Google's MorphNet does, we also implemented a momentum on the layer insertion where we skip the deletion of layers, if the accuracy we achieved with the new architecture is significantly higher than the accuracy of the previous architecture. Another feature is a learning rate on the MorphNet suggestions: We do not use these shrinkage suggestions exactly to get a better cooperation of the two basic structures MorphNet and LayerLasso.

Neural architecture search requires an order of magnitude more compute resources than an ordinary training. In addition, our architecture search maintains a number of hyper parameters such as different regularisation strengths and adjusting the momentum on layer insertion.
Therefore, the tuning of these hyper parameters consumes high amounts of compute resources.


A preliminary result on the CIFAR10 dataset is given in \cref{fig:CIFAR10_example}. We started with a minimal architecture containing only a single convolutional layer and then automatically added ResNet blocks with our architecture search, achieving a test accuracy of up to 92.97\% which is fairly close to comparable hand-crafted ResNet architectures.


\section{Conclusion \& Outlook}

In this overview article, we gave insight into our work on some of the most resource demanding applications in the field of deep learning and the role of uncertainty quantification therein.


For future work building upon our proposed anomaly segmentation method, our resulting model could be integrated in an retrieval-based approach to identify relevant OoD objects. Via clustering methods, the frequency of occurrence of specific OoD objects can then be assessed to determine whether new concepts are required to be learned. Ideally, this is pursued in an entirely unsupervised fashion, in this way enabling DNNs to operate in the open world \cite{Uhlemeyer2022TowardsUO, chan2022detecting}.
In the context of the domain adaptation we will investigate if an active learning strategy based on the uncertainty of the network can help sampling the most informative samples in a pool of unlabelled data which are then labelled and used for fine-tuning the generator. With a focus on active learning for object detection, current investigations show that query strategies which perform better than the random baseline in terms of queried images do not, however, consistently outperform the random baseline in terms of queried bounding boxes.
Since annotation costs are proportional to the amount of boxes queried, we plan on refining our method and investigating the problem of reducing the amount of labelled instances queried while maintaining performance via an informed selection strategy.
With regard to our automated design of neural networks, we will evaluate our methods on real-world data and will focus on optimising state-of-the-art networks for a number of benchmark datasets to further develop the method.

\bibliographystyle{nic}
\bibliography{biblio}

\section*{Acknowledgements}
\begin{wrapfigure}[10]{l}{0.17\linewidth}
  \begin{center}
    \vspace{-10pt}
    \includegraphics[width=\linewidth, trim=30pt 0 10pt 0]{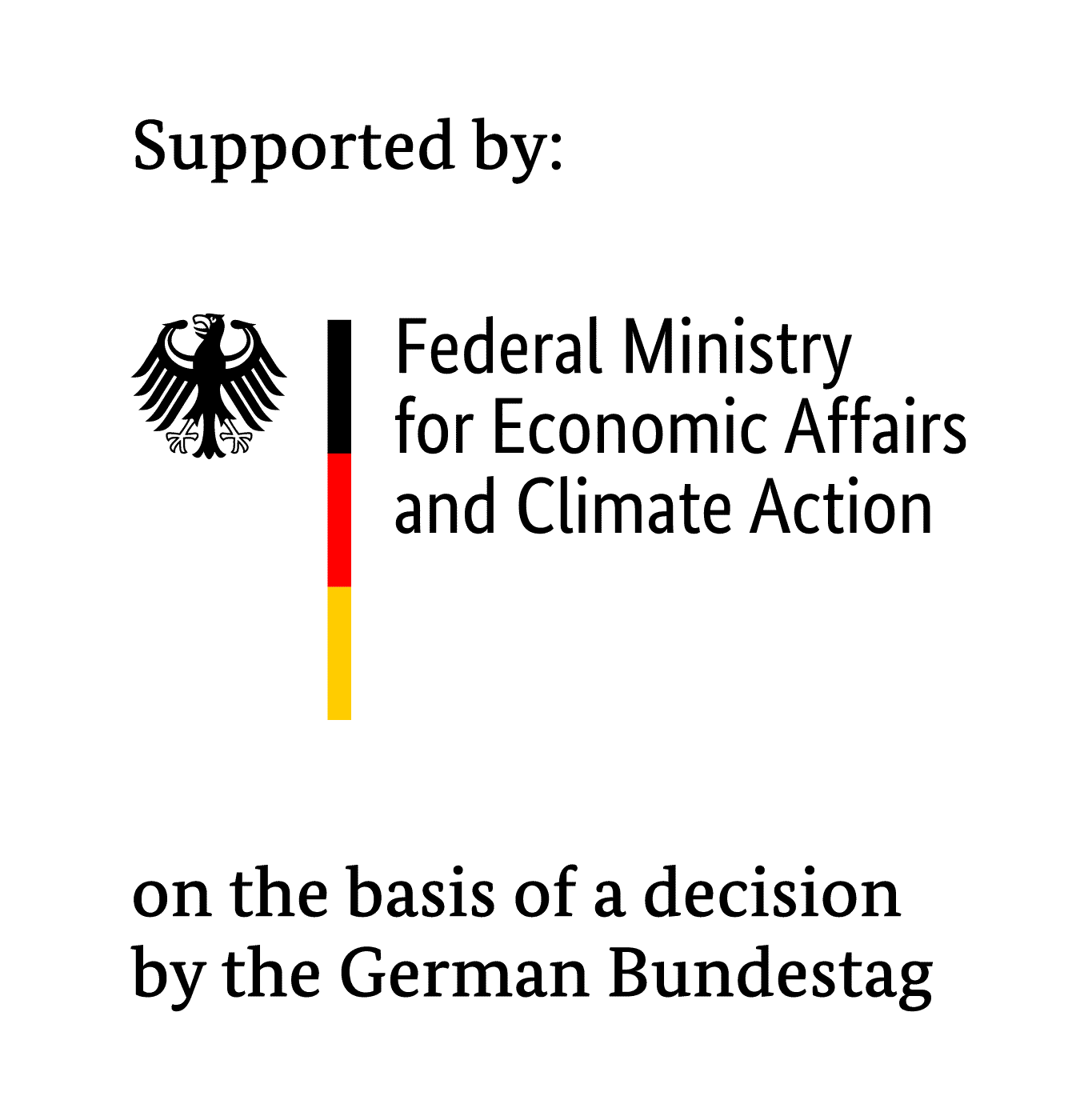}
    \includegraphics[width=\linewidth, trim=30pt 0 10pt 0]{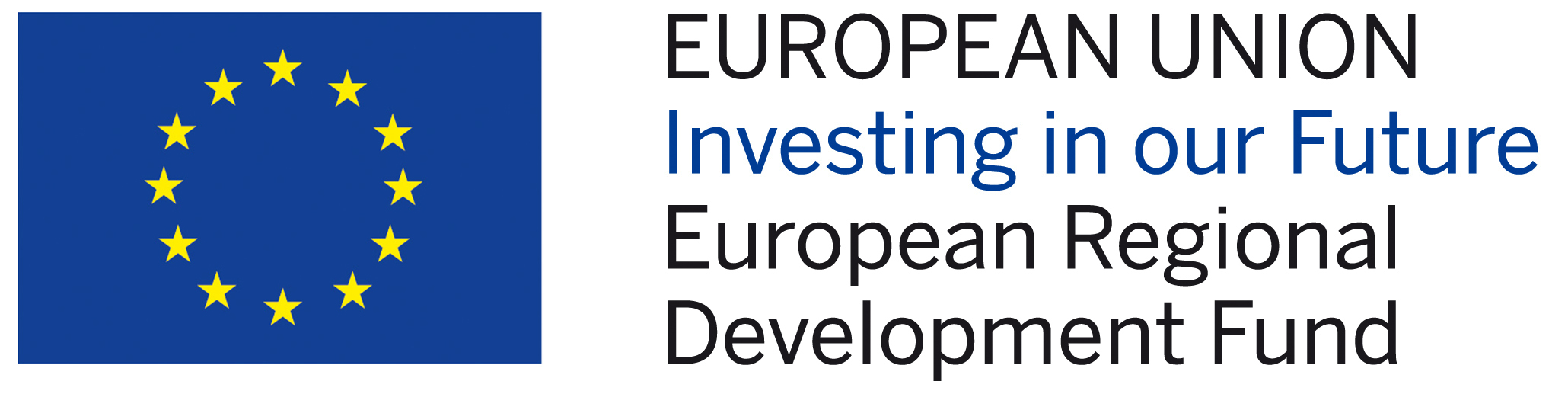}
  \end{center}
\end{wrapfigure}
The research leading to these results is in part funded by the German Federal Ministry for Economic Affairs and Climate Action within the project ``Methoden und Maßnahmen zur Absicherung von KI basierten Wahrnehmungsfunktionen für das automatisierte Fahren (KI-Absicherung)'', grant no.\ 19A19005R and ``KI Delta Learning - Scalable AI for Automated Driving'', grant no. 19A19013Q. Furthermore, the authors gratefully acknowledge financial support by the state Ministry of Economy, Innovation and Energy of Northrhine Westphalia (MWIDE) and the European Fund for Regional Development via the FIS.NRW project BIT-KI, grant no. EFRE-0400216. The authors would like to thank the consortiums for the successful cooperation.
Furthermore, the authors gratefully acknowledge the Gauss Centre for Supercomputing e.V. (www.gauss-centre.eu) for funding this project by providing computing time through the John von Neumann Institute for Computing (NIC) on the GCS Supercomputer JUWELS~\cite{JUWELS} at Jülich Supercomputing Centre (JSC).

\end{document}

%% file: images/meta_detect_pipeline.tex
\begin{tikzpicture}
    \pgfmathsetmacro{\cubex}{2}
    \pgfmathsetmacro{\cubey}{1.3}
    \pgfmathsetmacro{\cubez}{1.6}
    \pgfmathsetmacro{\topcentering}{-0.5*\cubey+0.3}
    \pgfmathsetmacro{\humanx}{0}
    \node[rectangle, rounded corners, fill=cyan!30, draw=cyan] (human) at (-0.5*\cubex+0.15,\topcentering-3.5) {Human Oracle \ManFace};

    \pgfmathsetmacro{\labeledx}{-4}
    \node[cylinder, draw, fill=green!30, shape border rotate=90, minimum width=1.2cm, minimum height=1cm, aspect=0.5] 
        (L) at (\labeledx, -2.2) {$\mathcal{L}$};

    \draw[blue,fill=blue!20] (0,0,0) -- ++(-\cubex,0,0) -- ++(0,-\cubey,0) -- ++(\cubex,0,0) -- cycle;
    \draw[blue,fill=blue!20] (0,0,0) -- ++(0,0,-\cubez) -- ++(0,-\cubey,0) -- ++(0,0,\cubez) -- cycle;
    \draw[blue,fill=blue!20] (0,0,0) -- ++(-\cubex,0,0) -- ++(0,0,-\cubez) -- ++(\cubex,0,0) -- cycle;
    \node[rectangle, minimum width=2.3cm, minimum height=2cm] (network) at (-0.5*\cubex+0.15,\topcentering) {};
    \node (networktxt) at (-0.5*\cubex,-0.5*\cubey) {DNN};

    \node[cylinder, draw, fill=green!30, shape border rotate=90, minimum width=1.2cm, minimum height=1cm, aspect=0.5] 
        (V) at (5, \topcentering) {$\mathcal{V}$};
    \node[cylinder, draw, fill=green!30, shape border rotate=90, minimum width=1.2cm, minimum height=1cm, aspect=0.5] 
        (T) at (5, \topcentering+1.8) {$\mathcal{T}$};
    \node[cylinder, draw, fill=green!30, shape border rotate=90, minimum width=1.2cm, minimum height=1cm, aspect=0.5] 
        (U) at (5, \topcentering-1.8) {$\mathcal{U}$};
    \node[draw] (map) at (8, \topcentering+1.8) {$\mathit{mAP}$};

    \node[draw, cylinder, shape border rotate=90, aspect=0.1, fill=red!20] 
        (metrics+gt) at (8, \topcentering) {\begin{tabular}{c}Metrics + \\ Ground Truth \end{tabular}};
    \node[draw, cylinder, shape border rotate=90, aspect=0.1, fill=red!20] 
        (metrics) at (8, \topcentering-1.8) {Metrics};
    \node[rectangle, rounded corners, fill=orange!30, draw=orange] 
        (metamodel) at (11.5, \topcentering) {Meta Model};
    \node[draw] (predunc) at (8.5, \topcentering-3.5) {\begin{tabular}{c}Predictive \\Uncertainty\end{tabular}};

    \node[draw] (imgunc) at (5, \topcentering-3.5) {\begin{tabular}{c}Image \\ Uncertainty\end{tabular}};

    \draw[-stealth, line width=1pt, out=90, in=180] (L) to node[midway, above, rotate=30]{train DNN}  (network);

    \draw[line width=1pt, out=0, in=180] (network) -+ node[midway, above]{predict} (3, \topcentering);
    \draw[-stealth, line width=1pt, out=90, in=180] (3, \topcentering) to (T);
    \draw[-stealth, line width=1pt, out=0, in=180] (3, \topcentering) to (V);
    \draw[-stealth, line width=1pt, out=270, in=180] (3, \topcentering) to (U);
    \draw[-, line width=1pt] (metamodel) to (11.5, \topcentering-1.8);
    \draw[-, line width=1pt] (metrics) to (11.5, \topcentering-1.8);
    \draw[-stealth, line width=1pt, out=270, in=0] (11.5, \topcentering-1.8) to node[midway, below, rotate=45]{predict} (predunc);

    \draw[-, line width=1pt] (T) to (map);
    \draw[-, line width=1pt] (V) to (metrics+gt);
    \draw[-, line width=1pt] (U) to (metrics);
    \draw[-stealth, line width=1pt] (metrics+gt) to node[midway, above]{fit} (metamodel);

    \draw[-, line width=1pt] (predunc) to (imgunc);
    \draw[-stealth, line width=1pt] (imgunc) -- node[midway]{\begin{tabular}{c}select query \\ $\mathcal{U}_\mathrm{select}$ \end{tabular}} (human);
    \draw[-stealth, line width=1pt, out=180, in=270] 
        (human) to node[midway, below, rotate=-35]{add $\mathcal{U}_\mathrm{select}$} (L);
    \draw[-stealth, dashed, draw=black!20, line width=1pt] (network) 
        to node[midway, rotate=270]{\textcolor{gray}{\begin{tabular}{c}random \\ query\end{tabular}}} (human);

\end{tikzpicture}

%% file: images/al_nutshell.tex
\begin{tikzpicture}[>=stealth]
\def\arcbegin{0}
\def\arcending{280}
\node (generation_bgd) at (3,8.2) {\includegraphics[width=3.9\textwidth]{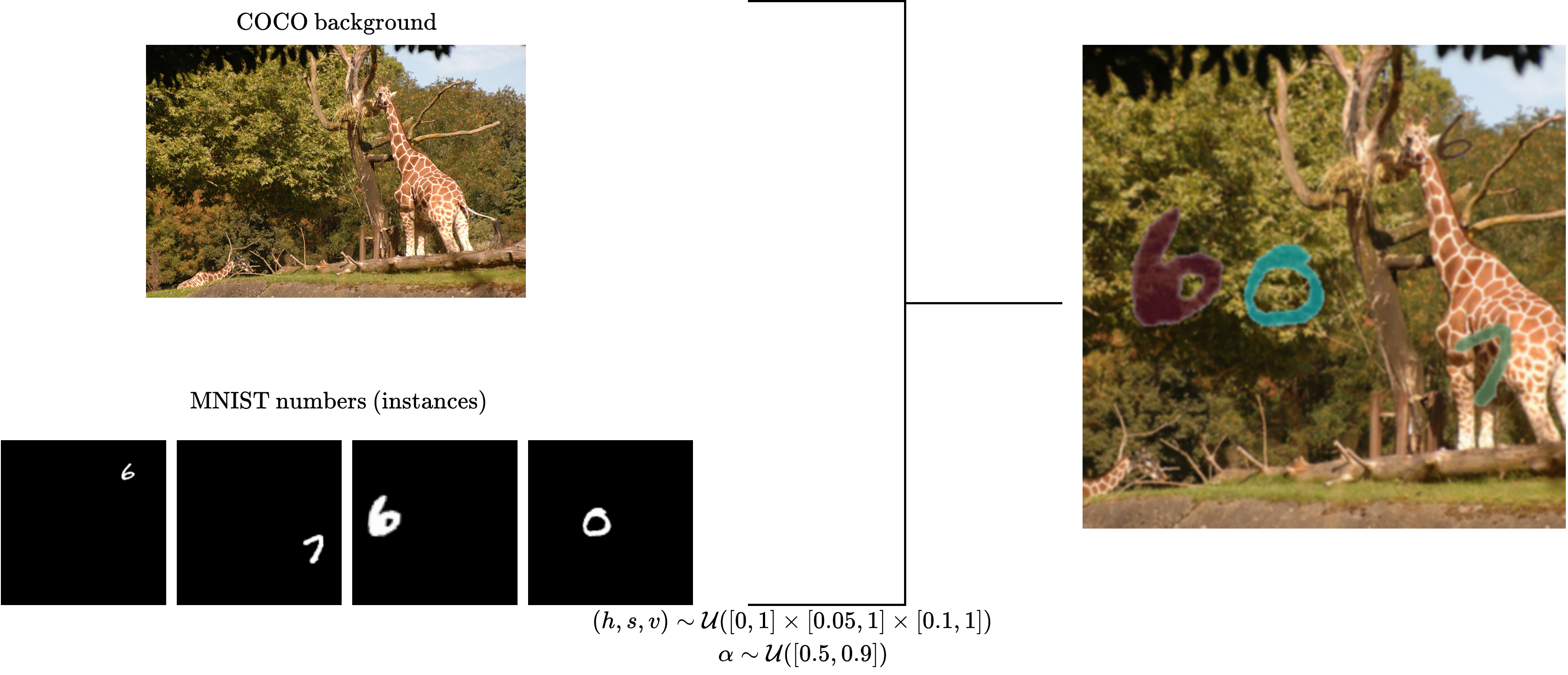}};
\begin{axis}[
    view={19}{30},
    axis lines=center,
    axis on top,
    domain=0:1,
    y domain=\arcbegin:\arcending,
    xmin=-1.5, xmax=1.5,
    ymin=-1.5, ymax=1.5,
    zmin=0.0, zmax = 1.2,
    hide axis,
    samples = 20,
    data cs=polar,
    mesh/color input=explicit mathparse,
    shader=interp]
\addplot3 [
    surf,
    variable=\u,
    variable y=\v,
    point meta={symbolic={Hsb=v,u,u}}] 
    (v,1,u);
\addplot3 [
    surf,
    samples = 50,
    variable=\u,
    variable y=\v,
    point meta={symbolic={Hsb=v,u,1}}] 
    (v,u,1);
\addplot3 [
    surf,
    variable=\u,
    y domain = 0:1,
    variable y=\w,
    point meta={symbolic={Hsb=\arcbegin,u,z}}]
    (\arcbegin,u,w);
\addplot3 [
    surf,
    variable=\u,
    y domain = 0:1,
    variable y=\w,
        point meta={symbolic={Hsb=\arcending,u,z}}]
    (\arcending,u,w);
\addplot3[
    line width=0.3pt]
    coordinates {(0,0,0) (\arcbegin, 1, 0) (\arcbegin,1,1) (0,0,1) ({(\arcending)},1,1) (\arcending, 1, 0) (0,0,0) };
\draw[
    line width = 0.3pt]
    (axis cs: {cos(\arcbegin)}, {sin(\arcbegin)},1) arc (\arcbegin:\arcending:1);
\draw[
    -,
    line width = 0.8pt]
    (axis cs: {1.5*cos(\arcbegin+20)}, {1.5*sin(\arcbegin+20)},0.5) arc ({\arcbegin}:{71.3}:1.5);
\draw[
    ->,
    line width = 0.8pt]
    (axis cs: {1.5*cos(\arcending-130)}, {1.5*sin(\arcending-130)},0.5) arc ({\arcending-130}:{\arcending}:1.5);
\addplot3[
    <->,
    line width=0.8pt]
    coordinates {(\arcbegin,1.5,0) (0,0,0) (0,0,1.5)};
\node at (axis cs:1.4,0,0) [anchor=north east] {S};
\node at (axis cs:-0.1,0,1.6) [anchor= north east] {V};
\node at (axis cs:-1.65,0.0,0.4) [anchor=east] {H};

\end{axis}

\end{tikzpicture}